\documentclass[onecolumn,sigconf]{acmart}
\usepackage[latin9]{inputenc}
\usepackage{array}
\usepackage{verbatim}
\usepackage{url}
\usepackage{multirow}
\usepackage{amsmath}
\usepackage{amssymb}
\usepackage{stmaryrd}
\usepackage{graphicx}

\makeatletter

\providecommand{\tabularnewline}{\\}

\usepackage{booktabs} 

\setcopyright{rightsretained}

\acmDOI{10.475/123_4}

\acmISBN{123-4567-24-567/08/06}

\acmConference[WOODSTOCK'97]{ACM Woodstock conference}{July 1997}{El
  Paso, Texas USA} 
\acmYear{1997}
\copyrightyear{2016}

\acmPrice{15.00}

\@ifundefined{showcaptionsetup}{}{%
 \PassOptionsToPackage{caption=false}{subfig}}
\usepackage{subfig}
\makeatother

\begin{document}

\title{PixelSNE: Visualizing Fast with Just Enough Precision \\
via Pixel-Aligned Stochastic Neighbor Embedding}
\titlenote{Produces the permission block, and   copyright information} 

\author{Minjeong Kim}  \affiliation{   \institution{Korea University}  } \email{mj1642@korea.ac.kr}
\author{Minsuk Choi}  \affiliation{   \institution{Korea University}  } \email{mchoi@korea.ac.kr}
\author{Sunwoong Lee}  \affiliation{   \institution{Korea University}  } \email{solomoj94@gmail.com}
\author{Jian Tang}  \affiliation{   \institution{University of Michigan}  } \email{jiant@umich.edu}
\author{Haesun Park}  \affiliation{   \institution{Georgia Tech}  } \email{hpark@cc.gatech.edu}
\author{Jaegul Choo}  \authornote{the corresponding author}  \affiliation{   \institution{Korea University}  } \email{jchoo@korea.ac.kr}

\renewcommand{\shortauthors}{Minjeong Kim et al.} 

\begin{abstract}
Embedding and visualizing large-scale high-dimensional data in a two-dimensional space is an important problem since such visualization can reveal deep insights out of complex data. Most of the existing embedding approaches, however, run on an excessively high precision, ignoring the fact that at the end, embedding outputs are mapped into coarse-grained pixel coordinates in a limited screen space. Motivated by this observation and directly considering it in an embedding algorithm, we accelerate Barnes-Hut tree-based t-distributed stochastic neighbor embedding (BH-SNE), known as a state-of-the-art 2D embedding method, and propose a novel alternative called PixelSNE, a highly-efficient, screen resolution-driven 2D embedding method with a linear computational complexity in terms of the number of data items. Our experimental results show the significantly fast running time of PixelSNE by a large margin against BH-SNE, while maintaining the comparable embedding quality. Finally, the source code of our method is publicly available at \url{https://github.com/awesome-davian/pixelsne}. 
\end{abstract}

\ccsdesc[500]{Computing methodologies~Feature selection}
\keywords{t-SNE; Barnes-Hut SNE; Dimension reduction; 2D embedding; Scatterplot; Visualization}
\maketitle

\section{Introduction}

\label{sec:intro} 

Visualizing high-dimensional data in a two-dimensional (2D) or three-dimensional
(3D)\footnote{This paper discusses only the 2D embedding case, but our proposed
approach can be easily extended to the 3D case. } scatterplot is an effective approach that can reveal deep insights
about data. Through such visualization, one can obtain the idea about
the relationships among clusters as well as those of individual data,
e.g., similar or outlying clusters and/or data items. 

To generate a scatterplot given high-dimensional data, one can apply
various dimension reduction or low-dimensional embedding approaches
including traditional methods (e.g., principal component analysis~\cite{jolliffe02pca}
and multidimensional scaling~\cite{kruskal64multidim,kruskal64numerical})
and recent manifold learning methods (e.g., isometric feature mapping~\cite{tenenbaum00global},
locally linear embedding~\cite{roweis03think}, and Laplacian Eigenmaps~\cite{belkin03laplacian}). 

These methods, however, do not properly handle the significant information
loss due to reducing high dimensionality down to two or three, and
in response, an advanced dimension reduction technique called t-distributed
stochastic neighbor embedding (t-SNE)~\cite{van08visualizing} has
been proposed, showing its outstanding advantages in generating 2D
scatterplots. A drawback of t-SNE is its significant computing time
against a large number of data items, e.g., the computational complexity
of $O\left(n^{2}\right)$, where $n$ represents the number of data
items. Although various approximation techniques attempting to accelerate
its algorithm have been proposed, e.g., Barnes-Hut SNE (BH-SNE)~\cite{van2014accelerating}
with the complexity of $\mathcal{O}\left(n\log n\right)$, it still
takes much time to apply them to large-scale data. 

To tackle this issue, this paper proposes the novel framework that
can significantly accelerate the 2D embedding algorithms in visualization
applications. The proposed framework is motivated by the fact that
most embedding approaches compute the low-dimensional coordinates
with an excessive precision, e.g., a double precision with 64-bit
floating point representation, compared to the limited precision that
our screen space has. For instance, if our screen space has $1024\times768$
pixels, then the resulting coordinates computed from an embedding
algorithm will be mapped with $1024$ and $768$ integer levels of
$x$ and $y$ coordinates, respectively. Moreover, when making sense
of a 2D scatterplot, human perception does not often require a high
precision from its results~\cite{choo13scr}. 

Leveraging this idea of the just enough precision for our screen and
human perception to the above-mentioned state-of-the-art method, BH-SNE,
we propose a significantly fast alternative called pixel-aligned SNE
(PixelSNE), which shows more than 5x fold speedup against BH-SNE for
421,161 data item of News aggregator dataset. In detail, by lowering
and matching the precision used in BH-SNE to that of pixels in a screen
space, PixelSNE directly optimizes %
2D-coordinates in the screen space with a pre-given resolution. 

In this paper, we describe the mathematical and algorithmic details
of how we utilized the idea of a pixel-based precision in BH-SNE and
present the extensive experimental results that show the significantly
fast running time of PixelSNE by a large margin against BH-SNE, while
maintaining the the embedding quality. 

Generally, our contributions can be summarized as follows: 

\noindent\textbf{1. }We present a novel framework of a pixel-based
precision driven by a screen space with a finite resolution. 

\noindent\textbf{2. }We propose a highly-efficient 2D embedding approach
called PixelSNE by leveraging our idea of a pixel-based precision
in BH-SNE. 

\noindent\textbf{3. }We perform extensive quantitative and qualitative
analyses using various real-world datasets, showing the significant
speedup of our proposed approach against BH-SNE along with a comparable
quality of visualization results.

\section{Related Work}

\label{sec:rel_work} 

Dimension reduction or low-dimensional embedding of high-dimensional
data~\cite{mateen09dimered} has long been an active research area.
Typically, most of these methods attempt to generate the low-dimensional
coordinates that maximally preserve the pairwise distances/similarities
given in a high-dimensional space. Such low-dimensional outputs generally
work for two purposes: (1) generating the new representations of original
data for improving the desired performance of a downstream target
task, such as its accuracy and/or computational efficiency, and (2)
visualizing high-dimensional data in a 2D scatterplot for providing
humans with the in-depth understanding and interpretation about data. 

Widely-used dimension reduction methods used for visualization application
include principal component analysis (PCA)~\cite{jolliffe02pca},
multidimensional scaling~\cite{kruskal64multidim,kruskal64numerical},
Sammon mapping~\cite{sammon69nonlinear}, generative topographic
mapping~\cite{bishop98gtm}, and self-organizing map~\cite{kohonen2001som}.
While these traditional methods generally focus on preserving global
relationships rather than local ones, a class of nonlinear, local
dimension reduction techniques called manifold learning~\cite{lee07nonlinear}
has been actively studied, trying to recover an intrinsic curvilinear
manifold out of given high-dimensional data. Representative methods
are isometric feature mapping~\cite{tenenbaum00global}, locally
linear embedding~\cite{roweis03think}, Laplacian Eigenmaps~\cite{belkin03laplacian},
maximum variance unfolding~\cite{weinberger06semidef}, and autoencoder~\cite{hinton06autoencoder}. 

Specifically focusing on visualization applications, a recent method,
t-distributed stochastic neighbor embedding~\cite{van08visualizing},
which is built upon stochastic neighbor embedding~\cite{hinton02sne},
has shown its superiority in generating the 2D scatterplots that can
reveal meaningful insights about data such as clusters and outliers.
Since then, numerous approaches have been proposed to improve the
visualization quality and its related performances in the 2D embedding
results. For example, a neural network has been integrated with t-SNE
to learn the parametric representation of 2D embedding~\cite{maaten09partsne}.
Rather than the Euclidean distance or its derived similarity information,
other information types such as non-metric similarities~\cite{maaten12nonmetric}
and relative ordering information about pairwise distances in the
form of similarity triplets~\cite{maaten12ste} have been considered
as the target information to preserve. Additionally, various other
optimization criteria and their optimization approaches, such as elastic
embedding~\cite{carreira2010elastic} and NeRV~\cite{venna10information},
have been proposed. 

The computational efficiency and scalability of 2D embedding approaches
has also been widely studied. An accelerated t-SNE based on the approximation
using the Barnes-Hut tree algorithm has been proposed~\cite{van2014accelerating}.
Gisbrecht et al. proposed a linear approximation of t-SNE~\cite{gisbrecht12lineartsne}.
More recently, an approximate, but user-steerable t-SNE, which provides
interactions with which a user can control the degree of approximation
on user-specified areas, has also been studied~\cite{pezzotti16atsne}.
In addition, a scalable 2D embedding technique called LargeVis~\cite{tang16largevis}
significantly reduced the computing times with a linear time complexity
in terms of the number of data items. 

Even with such a plethora of 2D embedding approaches, to the best
of our knowledge, none of the previous studies have directly exploited
the limited precision of our screen space and human perception for
developing highly efficient 2D embedding algorithms, and our novel
framework of controlling the precision in return for algorithm efficiency
and the proposed PixelSNE, which significantly improves the efficiency
of BH-SNE, can be one such example. 

\section{Preliminaries}

\label{sec:preliminaries} 

\begin{figure*}
\centering\includegraphics[clip,width=0.75\textwidth]{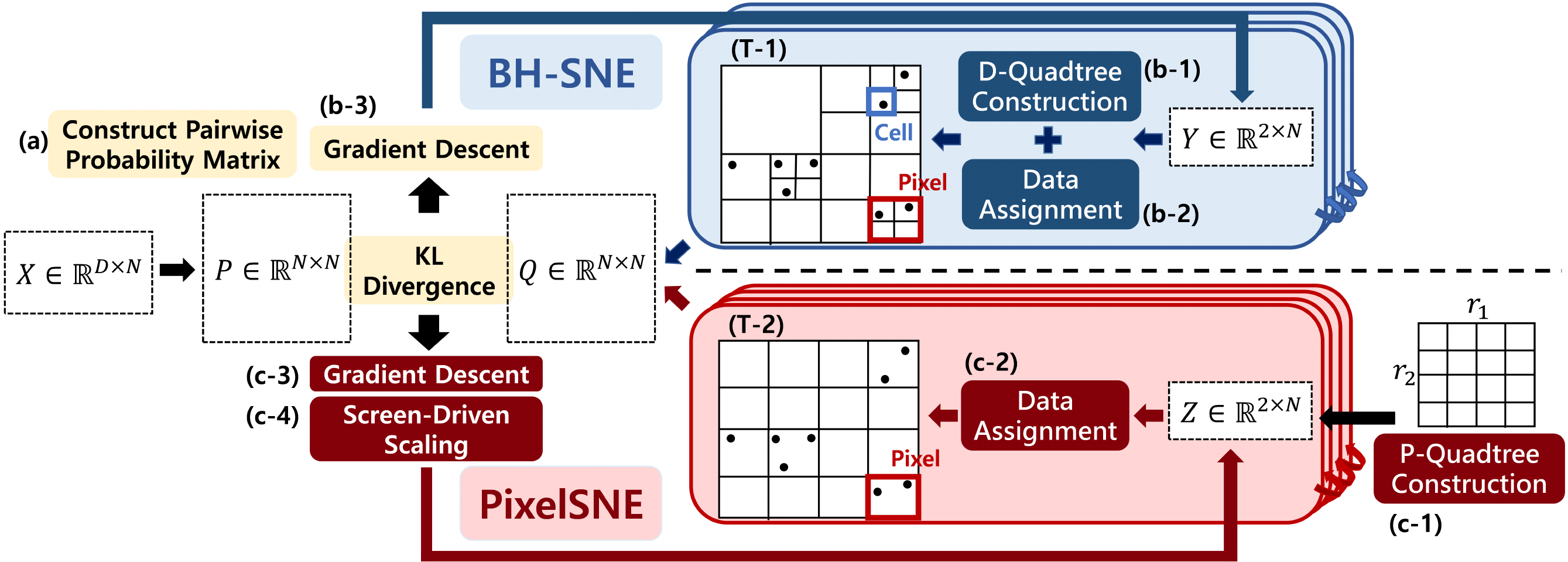}

\caption{Overview of the proposed PixelSNE in comparison with the original
Barnes-Hut SNE (BH-SNE)}
\label{fig:overview}
\end{figure*}

In this section, we introduce the problem formulation and the two
existing methods, t-distributed stochastic neighbor embedding (t-SNE)
and its efficient version called Barnes-Hut SNE (BH-SNE). 

\subsection{Problem Formulation}

A 2D embedding method takes a set of high-dimensional data items,
$\mathcal{X}=\{x_{i}\in\mathbf{\mathbb{R}}^{D}\}_{i=1,2,...,N}$,
where $N$ is the number of data items and $D$ is the original dimensionality,
and gives their 2D (low-dimensional) embedding, $\mathcal{Y}=\left\{ y_{i}\in\mathbb{R}^{2}\right\} _{i=1,2,..,N}$,
as an output. Given a screen of resolution $r_{1}\times r_{2}$, where
$r_{1}$ and $r_{2}$, the $x$- and $y$-axis resolutions, respectively,
are positive integers, the scatterplot is generated by transforming
$\mathcal{Y}$ to their corresponding (zero-based) pixel coordinates
\begin{equation}
\mathcal{Z}=\left\{ \left\lfloor z_{i}\right\rfloor :\,z_{i}=\left[\begin{array}{c}
z_{i,1}\\
z_{i,2}
\end{array}\right]\in\mathbb{R}^{2},\right\} _{i=1,2,..,N}\text{ where}\label{eq:z1}
\end{equation}
\begin{equation}
0\leq z_{i,d}<r_{d},\,\forall d\in\left\{ 1,\,2\right\} .\label{eq:z2}
\end{equation}

\subsection{t-Distributed Stochastic Neighbor Embedding (t-SNE)}

t-SNE embeds high-dimensional data into a low-dimensional space by
minimizing the differences between the joint probability distribution
representing pairwise relationships in $\mathcal{X}$ and its counterpart
in $\mathcal{Y}$. In detail, t-SNE computes the Euclidean pairwise
distance matrix $D_{\mathcal{X}}\in\mathbb{R}^{N\times N}$ of $\mathcal{X}$
and then converts it to the high-dimensional joint probability matrix
$P\in\mathbb{R}^{N\times N}$ using a Gaussian kernel. 

Next, given randomly initialized $\mathcal{Y}$, t-SNE computes the
Euclidean pairwise distance matrix $D_{\mathcal{Y}}\in\mathbb{R}^{N\times N}$
and then converts it to a low-dimensional joint probability matrix
$Q\in\mathbb{R}^{N\times N}$ using a Student's $t$-distribution.
To be specific, the $\left(i,\,j\right)$-th component $q_{ij}^{\left(t\right)}$
of $Q^{\left(t\right)}$ at iteration $t$, which represents the similarity
between $y_{i}^{\left(t\right)}$ and $y_{j}^{\left(t\right)}$ in
a probabilistic sense, is computed as 
\begin{equation}
q_{ij}^{\left(t\right)}=\left(\left(1+\left\Vert y_{i}^{\left(t\right)}-y_{j}^{\left(t\right)}\right\Vert _{2}^{2}\right)Z^{\left(t\right)}\right)^{-1}\label{eq:q}
\end{equation}
 where $Z^{\left(t\right)}=\sum_{k\neq l}\left(1+\left\Vert y_{i}^{\left(t\right)}-y_{j}^{\left(t\right)}\right\Vert _{2}^{2}\right)^{-1}$. 

The objective function of t-SNE is defined using the Kullback-Leibler
divergence between $P$ and $Q$ as
\begin{equation}
C=KL(P\bigparallel Q)=\sum_{i\neq j}p_{ij}log\frac{p_{ij}}{q_{ij}},\label{eq:cost_func}
\end{equation}
 and t-SNE iteratively performs gradient-decent update on $\mathcal{Y}$
where the gradient with respect to $y_{i}$ is computed~\cite{van2014accelerating}
as 
\begin{align}
\frac{\partial C}{\partial y_{i}^{\left(t\right)}}= & 4\left(F_{attr}+F_{rep}\right)=4\left(\sum_{j\neq i}p_{ij}q_{ij}^{\left(t\right)}Z^{\left(t\right)}\left(y_{i}^{\left(t\right)}-y_{j}^{\left(t\right)}\right)\right.\nonumber \\
 & \left.-\sum_{j\neq i}\left(q_{ij}^{\left(t\right)}\right)^{2}Z^{\left(t\right)}\left(y_{i}^{\left(t\right)}-y_{j}^{\left(t\right)}\right)\right).\label{eq:grad}
\end{align}
Note that every data point exerts an attraction and an repulsion forces
to one another based on the difference between the two pairwise joint
probability matrices $P$ and $Q$. 

\subsection{Barnes-Hut-SNE (BH-SNE)}

While t-SNE adopts a brute-force approach with the computational complexity
$\mathcal{O}\left(N^{2}\right)$ considering all the pairwise relationships,
BH-SNE adopts two different tree-based approximation methods to reduce
this complexity. The first one called the vantage-point tree~\cite{yianilos93vptree}
approximately computes $D_{\mathcal{X}}$ and then $P$ as sparse
matrices by ignoring small pairwise distances as zeros (Fig.~\ref{fig:overview}(a)).
This approximation reduces the complexity $\mathcal{O}\left(N^{2}\right)$
of computing $D_{\mathcal{X}}$ and $P$ to $\mathcal{O}\left(uN\log N\right)$
where $u$ is a pre-defined parameter of perplexity and $u\ll N$.
Accordingly, involving only those nonzero $p_{ij}$'s in the sparse
matrix $P$, the complexity of computing $F_{attr}$ in Eq.~(\ref{eq:grad})
reduces to $\mathcal{O}\left(uN\right)$. 

When it comes to optimizing low-dimensional coordinates (Fig.~\ref{fig:overview}(b)),
BH-SNE adopts Barnes-Hut algorithm~\cite{barnes86barneshut} to compute
$F_{rep}$ in Eq.~(\ref{eq:grad}). When $\bigparallel y_{i}-y_{j}\bigparallel\approx\bigparallel y_{i}-y_{k}\bigparallel\gg\bigparallel y_{j}-y_{k}\bigparallel$,
the forces of $y_{j}$ and $y_{k}$ to $y_{i}$ are similar to each
other when computing the gradient. Based on this observation, BH-SNE
uses Barnes-Hut algorithm to find a single representative point of
multiple data points used for gradient update. For example, if we
set the representative data point of $y_{j}$ and $y_{k}$ as $y_{s}$,
then the low-dimensional joint probability %
$q_{is}$ can be used to substitute each of $q_{ij}$ and $q_{ik}$. 

To dynamically obtain the representative points $y_{s}$'s, BH-SNE
constructs a quadtree at each iteration, recursively splitting the
2D region that contains $y_{i}$'s into four pieces. Each node, which
we call a cell $c$, contains the following information: 

\noindent\textbf{1. }the boundary $\mathcal{B}_{d}^{\left(c\right)}$
about its corresponding 2D region, i.e., 
\begin{equation}
\mathcal{B}_{d}^{\left(c\right)}=\left[b_{min,d}^{\left(c\right)},\,b_{max,d}^{\left(c\right)}\right)\text{ for }d=1,\,2\label{eq:boundary}
\end{equation}

\noindent\textbf{2. }the set $\mathcal{Y}_{c}$ of $y_{i}$'s contained
in $c$, i.e., 
\begin{equation}
\mathcal{Y}_{c}=\left\{ y_{i}:\,y_{i,d}\in\mathcal{B}_{d}^{\left(c\right)}\text{ for }d=1,\,2\right\} ,\label{eq:ycell}
\end{equation}
and its cardinality, $N_{c}=\left|\mathcal{Y}_{c}\right|$, and 

\noindent\textbf{3. }the center of mass, $y_{c}$, of $y_{i}$'s
in $c$, i.e., 
\begin{equation}
y_{c}=\frac{1}{N_{c}}\sum_{y_{i}\in\mathcal{Y}_{c}}y_{i}\label{eq:ctrofmass}
\end{equation}

Given $y_{i}^{\left(t\right)}$ at iteration $t$, BH-SNE starts by
forming a root node $c_{root}$, containing all the $y_{i}^{\left(t\right)}$'s,
by setting $\mathcal{Y}_{c_{root}}=\mathcal{Y}$ in Eq.~(\ref{eq:ycell})
and 
\begin{align}
b_{min,d}^{\left(c_{root}\right)} & =\min_{i}y_{i}^{\left(t\right)}\text{ and }\label{eq:bmin}\\
b_{max,d}^{\left(c_{root}\right)} & =\max_{i}y_{i}^{\left(t\right)}+\epsilon,\,\forall d\in\left\{ 1,\,2\right\} ,\label{eq:bmax}
\end{align}
where $\epsilon$ is a small number in Eq.~(\ref{eq:boundary}).
BH-SNE then recursively splits $c$ into four (equally-sized) quadrant
cells located at ``northwest'', ``northeast'', ``southwest'', and
``southeast'' by setting their boundaries accordingly, assigning $y^{\left(t\right)}$'s
to these child cells based on the boundary information in Eq.~(\ref{eq:ycell}),
and computing their centers of mass. As assigning $y^{\left(t\right)}$'s
one by one to the tree, the quadtree grows until each leaf node contains
at most a single $y_{i}^{\left(t\right)}$. 

Afterwards, when computing the gradient with respect to $y_{i}$ in
Eq.~(\ref{eq:grad}) (Fig.~\ref{fig:overview}(b-3)), BH-SNE traverses
the quadtree via depth-first search to determine whether $y_{c}$
can work as the representative point of those contained in $c$ based
on the criterion 
\begin{equation}
r_{c}/\bigl\Vert y_{i}-y_{c}\bigr\Vert_{2}^{2}<\theta,\label{eq:bh_criteria}
\end{equation}
where $r_{c}$ represents the diagonal length of the region of $c$
and $\theta$ is a threshold parameter. Once finding $y_{c}$ satisfying
this criterion, the term$-q_{ij}^{2}Z(y_{i}-y_{j})$ in Eq.~(\ref{eq:grad})
for those points contained in $c$ is approximated as $-N_{c}q_{i,c}^{2}Z(y_{i}-y_{c})$,
thus reducing the computational complexity of  $F_{rep}$ in Eq.~(\ref{eq:grad})
to $\mathcal{O}(N\log N)$.

\section{Pixel-Aligned SNE (PixelSNE)}

\label{sec:sasne} 

In this section, we present our PixelSNE, which significantly reduces
the computational time of BH-SNE. 

\subsection{Pixel-Aligned Barnes-Hut Tree}

A major novelty of PixelSNE originates from the fact that it directly
considers the screen-space coordinates $z_{i}$ (Eq.~(\ref{eq:z1}))
instead of $y_{i}$. That is, PixelSNE utilizes the two properties
of $z_{i}$ during its optimization process: (1) the range of $z_{i,d}$
remains fixed as $\left[0,\,r_{d}\right)$ for $d=1,\,2$ throughout
algorithm iterations (Eq.~(\ref{eq:z2})) and (2) the precision of
$z_{i}$ is limited by the screen space resolution. %
{} Utilizing these characteristics, we accelerate the Barnes-Hut tree
construction as the assignment process of $z_{i}$'s to cells as follows.
For clarification, we denote our accelerated quadtree algorithm of
PixelSNE as a pixel-aligned quadtree (P-Quadtree) while we call the
original quadtree algorithm of BH-SNE as a data-driven quadtree (D-Quadtree). 

\noindent\textbf{One-time tree construction (instead of every iteration).}
Unlike BH-SNE, which builds a quadtree from scratch per iteration,
the above-mentioned properties of $z_{i}$ allow PixelSNE to build
P-Quadtree just one time before the main iterations of gradient-decent
update and to recycle it during the iterations. That is, PixelSNE
pre-computes the full information about (1) the boundary (Eq.~(\ref{eq:boundary}))
of each cell and (2) its center of mass (Eq.~(\ref{eq:ctrofmass}))
as follows. 

For the boundary information, we utilize the fact that since $\min_{i}z_{i,d}^{\left(t\right)}=0$
and $\max_{i}z_{i,d}^{\left(t\right)}=r_{d}$ for every iteration,
Eqs.~(\ref{eq:bmin}) and (\ref{eq:bmax}) boil down to 
\begin{align*}
b_{min,d}^{\left(c_{root}\right)} & =0\text{ and }\\
b_{max,d}^{\left(c_{root}\right)} & =r_{d}+\epsilon,\,\forall d\in\left\{ 1,\,2\right\} ,
\end{align*}
 which is no longer dependent on iteration $t$. This constant boundary
of the root node makes those of all the child cells in P-Quadtree
constant as well. This results a fixed depth of P-Quadtree as long
as the minimum size of the cell is determined, which will be discussed
later part of this sub-section in detail. Based on this idea, PixelSNE
pre-computes the boundaries of all the cells in P-Quadtree. 

Next, since the boundary of each cell is already determined, we simply
approximate the center of mass $y_{c}$ as the center location of
the cell corresponding region, e.g., 
\[
y_{c,d}=\frac{b_{min,d}^{\left(c\right)}+b_{max,d}^{\left(c\right)}}{2},\,\forall d\in\left\{ 1,\,2\right\} ,
\]
which is also not dependent on iteration $t$. 

Once the pre-computation of the above information (Fig.~\ref{fig:overview}(c-1))
finishes, the iterative gradient optimization in PixelSNE simply assigns
$z_{i}^{(t)}$'s to these pre-computed cells in P-Quadtree and updates
$\mathcal{Y}_{c}$ and $N_{c}$ (Fig.~\ref{fig:overview}(c-2)).
Note that in contrast, BH-SNE iteratively computes all these steps
every iteration (Figs.~\ref{fig:overview}(b-1) and (b-2)), which
acts as the critical inefficiency compared to PixelSNE.

\noindent\textbf{Bounded tree depth based on screen resolution.}
Considering a typical screen resolution, BH-SNE performs an excessively
precise computation. That is, when mapping D-Quadtree to pixel grids
of our screen, one can view that BH-SNE subdivides the pixels, the
minimum unit of the screen, into much smaller cells until each left
cell contains at most one data point (Fig.~\ref{fig:overview}(T-1)).

On the contrary, the minimum size of the cell in P-Quadtree is bounded
to the pixel size to avoid an excessive precision in computation (Fig.~\ref{fig:overview}(T-2)).
In detail, P-Quadtree keeps splitting the cell $c$ while satisfying
the condition, 
\[
b_{max,d}^{\left(c\right)}-b_{min,d}^{\left(c\right)}>1,\,\exists d\in\left\{ 1,\,2\right\} ,
\]
which indicates that the cell length is larger than the unit pixel
size $1$ for at least one of the two axes. As a result, the depth
of P-Quadtree is bounded by 
\[
\max_{d\in\left\{ 1,\,2\right\} }\left\lfloor \log_{2}r_{d}\right\rfloor .
\]

Such a bounded depth of P-Quadtree causes a leaf node to possibly
contain multiple data points, and in the gradient-descent update step,
we may not find the cell satisfying Eq.~(\ref{eq:bh_criteria}) even
after reaching a leaf node in the depth-first search traversal. In
this case, we just stop the traversal and use the center of mass of
the corresponding leaf node to approximately compute $F_{rep}$ (Eq.~(\ref{eq:grad}))
of those points contained in the node. 

Finally, let us clarify that we maintain $z_{i}$'s as double-precision
numbers just as BH-SNE, but our idea of a limited precision is realized
mainly by the two above-described novel techniques. 

\noindent\textbf{Computational complexity.} The depth of the Barnes-Hut
tree acts as a critical factor in algorithm efficiency since both
the assignment of data point to cells and the approximation of $F_{rep}$
(Eq.~(\ref{eq:grad})) are performed based depth-first search traversal,
the computational complexity of which is linear to the tree depth.
{} In the worst case where a majority of data points are located in
a relatively small region in a 2D space and a few outliers are located
far from them, D-Quadtree can grow as deeply as the depth of $N$,
which is much larger than that of P-Quadtree, $\max_{d\in\left\{ 1,\,2\right\} }\left\lfloor \log_{2}r_{d}\right\rfloor $,
when $r_{d}\ll N,\,\forall d\in\left\{ 1,\,2\right\} $. Owing to
this characteristics, the overall computational complexity of PixelSNE
is obtained as $\mathcal{O}\left(N\times\max_{d\in\left\{ 1,\,2\right\} }\left\lfloor \log_{2}r_{d}\right\rfloor \right)$,
which reduces to the linear computational complexity in terms of the
number of data items, given the fixed screen resolution $r_{d}$'s. 

Intuitively, BH-SNE traverses much finer-grained nodes to obtain the
excessively precise differentiation among data points. On the other
hand, PixelSNE assumes that as long as multiple data points are captured
within a single pixel, they are close enough to be represented as
the center of mass for their visualization in a screen. This guarantees
the depth of P-Quadtree to be limited by the screen resolution, instead
of being influenced by the undesirable non-uniformity of the 2D embedding
results during algorithm iterations. 

\subsection{Screen-Driven Scaling }

In order for the outputs $z_{i}$'s at every iteration to satisfy
Eq.~(\ref{eq:z2}) after their gradient-descent update, we scale
them as follows. Let us denote $z_{i}^{(t-1)}$ as the 2D coordinates
computed at iteration $t-1$ and $\hat{z}_{i}^{(t)}$ as its updated
coordinates at the next iteration $t$ via a gradient-descent method.
Note that $z_{i}^{(t-1)}$ satisfies Eq.~(\ref{eq:z2}) while $\hat{z}_{i}^{(t)}$
does not. Hence, we normalize each of the 2D coordinates of $\hat{z}_{i}^{(t)}$
and obtain $z_{i}^{(t)}=\left[\begin{array}{c}
z_{i,1}\\
z_{i,2}
\end{array}\right]$ as 
\[
z_{i,d}^{(t)}=\frac{r_{d}\left(\hat{z}_{i,d}^{(t)}-\min_{i}\hat{z}_{i,d}^{(t)}\right)}{\gamma_{d}^{(t)}},\,\forall d\in\left\{ 1,\,2\right\} ,
\]
 where $\gamma_{d}^{(t)}=\max_{i}\hat{z}_{i,d}^{(t)}-\min_{i}\hat{z}_{i,d}^{(t)}+\epsilon$
{} and $\epsilon$ is a small constant, e.g., $10^{-6}$. The reason
for introducing $\epsilon$ is to have the integer-valued 2D pixel
coordinates $\left\lfloor z_{i}\right\rfloor $ in Eq.~(\ref{eq:z1})
lie exactly between $0$ and $\left(r_{d}-1\right)$ for $d=1,\,2$.
For example, for a $1024\times768$ resolution, we impose the pixel
coordinates at each iteration to exactly have the range from $0$
to $1023$ and that from $0$ to $767$ for $x$- and $y$-coordinates,
respectively. %

This scaling step, however, affects the computation of the low-dimensional
pairwise probability matrix $Q$ since it scales each pairwise Euclidean
distance, $\left\Vert y_{i}-y_{j}\right\Vert $, in Eq.~(\ref{eq:q}),
resulting in a different probability distribution from the case with
no scaling. To compensate this scaling effect in computing $Q$, we
re-define $Q$ with respect to the scaled $z_{i}$'s as 
\begin{equation}
q_{ij}^{\left(t\right)}=\left(\left(1+\sum_{d=1}^{2}\beta_{d}^{\left(t\right)}\left(z_{i,d}^{\left(t-1\right)}-z_{j,d}^{\left(t-1\right)}\right)^{2}\right)Z^{\left(t\right)}\right)^{-1}\label{eq:q_wrt_z}
\end{equation}
 where $Z^{\left(t\right)}=\sum_{k\neq l}\left(1+\sum_{d=1}^{2}\beta_{d}^{\left(t\right)}\left(z_{i,d}^{\left(t-1\right)}-z_{j,d}^{\left(t-1\right)}\right)^{2}\right)^{-1}$
and $\beta_{d}$ is defined as 
\begin{equation}
\beta_{d}^{\left(t\right)}=\prod_{s=1}^{t-1}\left(\frac{\gamma_{d}^{(t)}}{r_{d}}\right)^{2},\,\forall d\in\left\{ 1,\,2\right\} .\label{eq:beta_scaling}
\end{equation}
where $\gamma_{d}^{(1)}=\epsilon$, since $\hat{z}_{i}^{(1)}$ is
randomly initialized from $\mathcal{N}(0,1)$. By introducing $\beta_{d}^{\left(t\right)}$
defined in this manner, PixelSNE uses Eq.~(\ref{eq:q_wrt_z}), which
is still equivalent to Eq.~(\ref{eq:q}). 

\subsection{Accelerating Construction of $P$}

\label{subsec:constructing_P}

To accelerate the process of constructing the matrix $P$, we adopt
a recently proposed, highly efficient algorithm of constructing the
approximate $k$-nearest neighbor graph (K-NNG)~\cite{tang16largevis}.
This algorithm builds on top of the classical state-of-the-art algorithm
based on random-projection trees but significantly improves its efficiency.
It first starts with a few random projection trees to construct an
approximate K-NNG. Afterwards, based on the intuition that ``the neighbors
of my neighbors are also likely to be my neighbors,'' the algorithm
iteratively improves the accuracy of K-NNG by exploring the neighbors
of neighbors defined according to the current K-NNG. In our experiments,
we show that this algorithm is indeed able to significantly accelerate
the process of constructing $P$.

\subsection{Implementation Improvements }

\label{subsec:minor_improvements}

The implementation of PixelSNE is built upon that of BH-SNE publicly
available at \url{https://github.com/lvdmaaten/bhtsne}. Other than
the above algorithmic improvements, we made implementation improvements
as follows. 

First, the Barnes-Hut tree involves many computations of division
by two (or powers of two). We replaced such computations by more efficient,
bitwise left-shift operations. Similarly, we replaced modulo-two operations,
which is used in data point assignment to cells, with bitwise masking
with $\mathtt{0x0001}$. Finally, we replaced the `pow' function from
`math' library in C/C++, which is used for squared-distance computation
with a single multiplication, e.g., $x\times x$ instead of `pow($x$,~2)'.
Although not significant, these implementation modifications rendered
consistently faster computing times than the case without them. 

\section{Experiments}

\label{sec:experiments} 

In this section, we present the quantitative analyses as well as qualitative
visualization examples of PixelSNE in comparison with original t-SNE
and BH-SNE. All the experiments were conducted on a single desktop
computer with Intel Core i7-6700K processors, 32GB RAM, and Ubuntu
16.04.1 LTS installed. 

\begin{table*}[th]
\caption{Comparison of computing times among 2D embedding algorithms. The t-SNE
results are excluded for large datasets due to its long computing
time. The total computing times are shown in boldface, and the numbers
in parentheses represent the standard deviation from three repetitions. }
\label{tab: total_compute_time}

\centering%
\begin{tabular}{|c||c|c|c|c|c|c|c|c|c|c|c|c|c|c|c|}
\hline 
\multirow{2}{*}{} & \multicolumn{3}{c|}{\textbf{\small{}20News}} & \multicolumn{3}{c|}{\textbf{\small{}FacExp}} & \multicolumn{3}{c|}{\textbf{\small{}MNIST}} & \multicolumn{3}{c|}{\textbf{\small{}NewsAgg}} & \multicolumn{3}{c|}{\textbf{\small{}Yelp}}\tabularnewline
\cline{2-16} 
 & \textit{\footnotesize{}P} & {\footnotesize{}Coord} & {\footnotesize{}Total} & \textit{\footnotesize{}P} & {\footnotesize{}Coord} & {\footnotesize{}Total} & \textit{\footnotesize{}P} & {\footnotesize{}Coord} & {\footnotesize{}Total} & \textit{\footnotesize{}P} & {\footnotesize{}Coord} & {\footnotesize{}Total} & \textit{\footnotesize{}P} & {\footnotesize{}Coord} & {\footnotesize{}Total}\tabularnewline
\hline 
\hline 
\multirow{2}{*}{{\small{}t-SNE}} & {\scriptsize{}2.43m} & {\scriptsize{}36.87m} & \textbf{\scriptsize{}39.31m} & {\scriptsize{}6.48m} & {\scriptsize{}84.94m} & \textbf{\scriptsize{}91.41m} & \multirow{2}{*}{{\scriptsize{}(-)}} & \multirow{2}{*}{{\scriptsize{}(-)}} & \multirow{2}{*}{{\scriptsize{}(-)}} & \multirow{2}{*}{{\scriptsize{}(-)}} & \multirow{2}{*}{{\scriptsize{}(-)}} & \multirow{2}{*}{{\scriptsize{}(-)}} & \multirow{2}{*}{{\scriptsize{}(-)}} & \multirow{2}{*}{{\scriptsize{}(-)}} & \multirow{2}{*}{{\scriptsize{}(-)}}\tabularnewline
 & {\scriptsize{}(0.98s)} & {\scriptsize{}(55.08s)} & {\scriptsize{}(56.06s)} & {\scriptsize{}(0.23s)} & {\scriptsize{}(10.12s)} & {\scriptsize{}(10.13s)} &  &  &  &  &  &  &  &  & \tabularnewline
\hline 
\multirow{2}{*}{{\small{}BH-SNE}} & {\scriptsize{}22.25s} & {\scriptsize{}156.70s} & \textbf{\scriptsize{}178.95s} & {\scriptsize{}0.19m} & {\scriptsize{}5.48m} & \textbf{\scriptsize{}5.56m} & {\scriptsize{}5.10m} & {\scriptsize{}13.93m} & \textbf{\scriptsize{}19.03m} & {\scriptsize{}3.96h} & {\scriptsize{}1.81h} & \textbf{\scriptsize{}5.77h} & {\scriptsize{}12.44h} & {\scriptsize{}5.83h} & \textbf{\scriptsize{}18.27h}\tabularnewline
 & {\scriptsize{}(1.55s)} & {\scriptsize{}(6.62s)} & {\scriptsize{}(8.16s)} & {\scriptsize{}(0.84s)} & {\scriptsize{}(40.73s)} & {\scriptsize{}(39.89s)} & {\scriptsize{}(10.84s)} & {\scriptsize{}(2.27m)} & {\scriptsize{}(2.45m)} & {\scriptsize{}(7.05m)} & {\scriptsize{}(4.04m)} & {\scriptsize{}(11.45m)} & {\scriptsize{}(19.29m)} & {\scriptsize{}(17.93m)} & {\scriptsize{}(37.22m)}\tabularnewline
\hline 
\multirow{2}{*}{{\small{}PixelSNE-VP}} & {\scriptsize{}14.68s} & {\scriptsize{}70.71s} & \textbf{\scriptsize{}85.39s} & {\scriptsize{}0.14m} & {\scriptsize{}2.04m} & \textbf{\scriptsize{}2.19m} & {\scriptsize{}3.76m} & {\scriptsize{}6.17m} & \textbf{\scriptsize{}9.92m} & {\scriptsize{}2.84h} & {\scriptsize{}0.63h} & \textbf{\scriptsize{}3.46h} & {\scriptsize{}8.89h} & {\scriptsize{}3.11h} & \textbf{\scriptsize{}12.00h}\tabularnewline
 & {\scriptsize{}(1.46s)} & {\scriptsize{}(1.58s)} & {\scriptsize{}(3.03s)} & {\scriptsize{}(0.66s)} & {\scriptsize{}(3.52s)} & {\scriptsize{}(4.16s)} & {\scriptsize{}(16.48s)} & {\scriptsize{}(11.92s)} & {\scriptsize{}(28.37s)} & {\scriptsize{}(5.49m)} & {\scriptsize{}(1.04m)} & {\scriptsize{}(6.52m)} & {\scriptsize{}(14.26m)} & {\scriptsize{}(7.47m)} & {\scriptsize{}(21.71m)}\tabularnewline
\hline 
\multirow{2}{*}{{\small{}PixelSNE-RP}} & {\scriptsize{}15.53s} & {\scriptsize{}72.17s} & \textbf{\scriptsize{}87.70s} & {\scriptsize{}0.32m} & {\scriptsize{}2.04m} & \textbf{\scriptsize{}2.36m} & {\scriptsize{}1.40m} & {\scriptsize{}6.05m} & \textbf{\scriptsize{}7.45m} & {\scriptsize{}0.30h} & {\scriptsize{}0.63h} & \textbf{\scriptsize{}0.94h} & {\scriptsize{}0.76h} & {\scriptsize{}3.14h} & \textbf{\scriptsize{}3.90h}\tabularnewline
 & {\scriptsize{}(0.49s)} & {\scriptsize{}(1.44s)} & {\scriptsize{}(1.93s)} & {\scriptsize{}(1.44s)} & {\scriptsize{}(3.57s)} & {\scriptsize{}(4.99s)} & {\scriptsize{}(2.22s)} & {\scriptsize{}(7.81s)} & {\scriptsize{}(9.86s)} & {\scriptsize{}(0.43m)} & {\scriptsize{}(2.08m)} & {\scriptsize{}(2.50m)} & {\scriptsize{}(0.97m)} & {\scriptsize{}(10.23m)} & {\scriptsize{}(11.20m)}\tabularnewline
\hline 
\end{tabular}
\end{table*}

\subsection{Experimental Setup}

This section describes our experimental setup including datasets and
compared methods. 

\subsubsection{Datasets}

Our experiments used four real-world datasets as follows: (1) \textbf{MNIST}
digit images, (2) Facial expression images (\textbf{FaceExp}), (3)
20 Newsgroups documents (\textbf{20News}), (4) News aggregator dataset
(\textbf{NewsAgg)} and (5) \textbf{Yelp} reviews.

\noindent\textbf{MNIST }\footnote{\url{http://cs.nyu.edu/~roweis/data.html}}
dataset contains a set of 70,000 gray-scale handwritten digit images
with $28\times28=784$ pixels, resulting in an input matrix $\mathcal{X}\in\mathbb{R}^{784\times70,000}$.
Each image has its digit label among ten different digits. 

\noindent\textbf{FacExp} \footnote{\url{https://goo.gl/W3Z8qZ}}
dataset contains a set of 28,709 gray-scale facial image data with
$48\times48$ pixels, with the label information out of seven different
facial expression categories. For this dataset, we extracted the features
from each image using a convolutional neural network. This network
starts with four convolutional layers with the 32, 64, 96, and 128
filters, respectively, with the size of each filter being $3\times3$.
Each convolutional layer is followed by the batch normalization layer,
rectified linear unit (ReLU), and the max-pooling layer over $2\times2$
patches. After then, two fully connected layers follow: the first
one with 256 output nodes followed by ReLU and the second with 256
output nodes followed by a softmax classifier for seven facial expression
categories. After training this model, we used the last hidden layer
output of 256 dimensions as the feature vector of each image, resulting
in an input matrix $\mathcal{X}\in\mathbb{R}^{256\times28,709}$. 

\noindent\textbf{20News }\footnote{\url{http://qwone.com/~jason/20Newsgroups/}}
dataset is a collection of 18,846 newsgroup posts on 20 different
newsgroup categories. We used the label of each document as one of
seven higher-level newsgroup categories. From each document, we extracted
the embedding vectors of those words in it using a pre-trained word2vec
embedding vector\footnote{\url{https://github.com/mmihaltz/word2vec-GoogleNews-vectors}}
and then averaged them to represent a single document, resulting in
an input matrix $\mathcal{X}\in\mathbb{R}^{300\times18,846}$. 

\noindent\textbf{NewsAgg }\footnote{\url{https://www.kaggle.com/uciml/news-aggregator-dataset} . }
dataset contains a collection of 421,161 news headlines on four different
categories. We represent the each headline with the averaged words
vector pre-processed as explained above. This results in an input
matrix $\mathcal{X}\in\mathbb{R}^{300\times421,161}$. 

\noindent\textbf{Yelp }\footnote{\url{https://www.yelp.com/dataset_challenge}}
review dataset contains around 2.7 million reviews written by around
687,000 users. We pre-processed this dataset by removing those rare
keywords appearing in less than 20 documents and other general keywords
such as food, restaurant, etc. After forming a term-document matrix,
we randomly selected one million reviews from it, resulting in an
input matrix $\mathcal{X}\in\mathbb{R}^{33,706\times1,000,000}$.
Afterwards, we extracted topic labels out of ten topics computed by
nonnegative matrix factorization~\cite{lee99learning} as a topic
modeling method. 

For all datasets, we reduced the dimensionality, i.e.g, the number
of rows of each input matrix to 50 by applying PCA, which is a standard
pre-processing step of t-SNE and BH-SNE. %

\begin{figure}
\centering\includegraphics[clip,width=1\columnwidth]{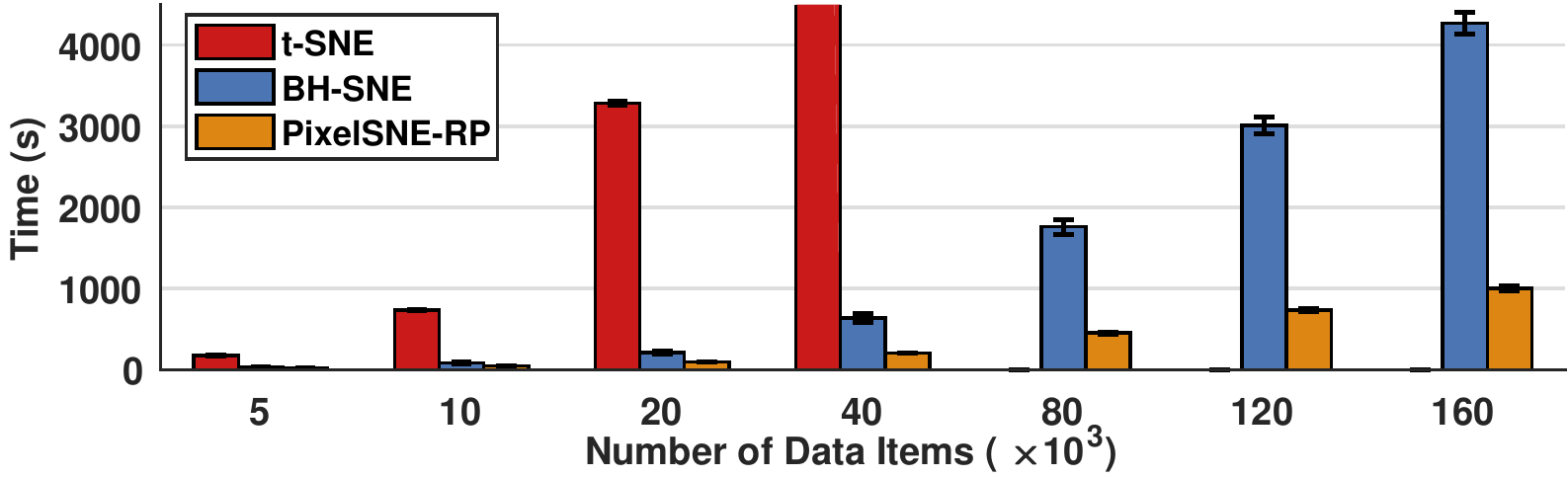}

\caption{Computing times vs.~the number of data items on NewsAgg dataset.
The error bar represents the standard deviation from five repetitions.
The t-SNE results are excluded for large datasets due to its long
computing time. }
\label{fig:Total_time}\vspace{-0in}
\end{figure}

\subsubsection{Compared Methods }

We compared our methods against the original t-SNE and BH-SNE. For
both methods, we used the publicly available code written by the original
author.\footnote{\url{https://lvdmaaten.github.io/tsne/}} We used
the default parameter values for both methods, e.g., the perplexity
value as 50, the number of iterations as 1,000, and the threshold
$\theta$ in Eq.~(\ref{eq:bh_criteria}) as 0.5. 

For our PixelSNE, we used its two different versions depending on
the algorithm for constructing $P$: (1) the vantage-point tree (\textbf{PixelSNE-VP})
used originally in BH-SNE and (2) the random projection tree-based
approach (\textbf{PixelSNE-RP}) used in LargeVis~\cite{tang16largevis}
(Section~\ref{subsec:constructing_P}). For the latter, we extracted
the corresponding part called $k$-nearest neighbor construction from
the publicly available code\footnote{\url{https://github.com/lferry007/LargeVis}}
and integrated it with our implementation of PixelSNE with its default
parameter settings except for setting the number of threads as $1$
to exclude the effect of parallelism. We set the number of iterations
and the threshold $\theta$ in Eq.~(\ref{eq:bh_criteria}) as the
same as BH-SNE, e.g., 1,000 and 0.5, respectively. We set both of
the screen resolution parameters $r_{1}$ and $r_{2}$ as 512 for
20News and FecExp datasets. For MNIST, NewsAgg and Yelp datasets,
both are set as 1024, 2048 and 8192, respectively. 

\begin{figure}[t]
\centering\subfloat[30,000 data items]{\centering\includegraphics[width=0.5\columnwidth]{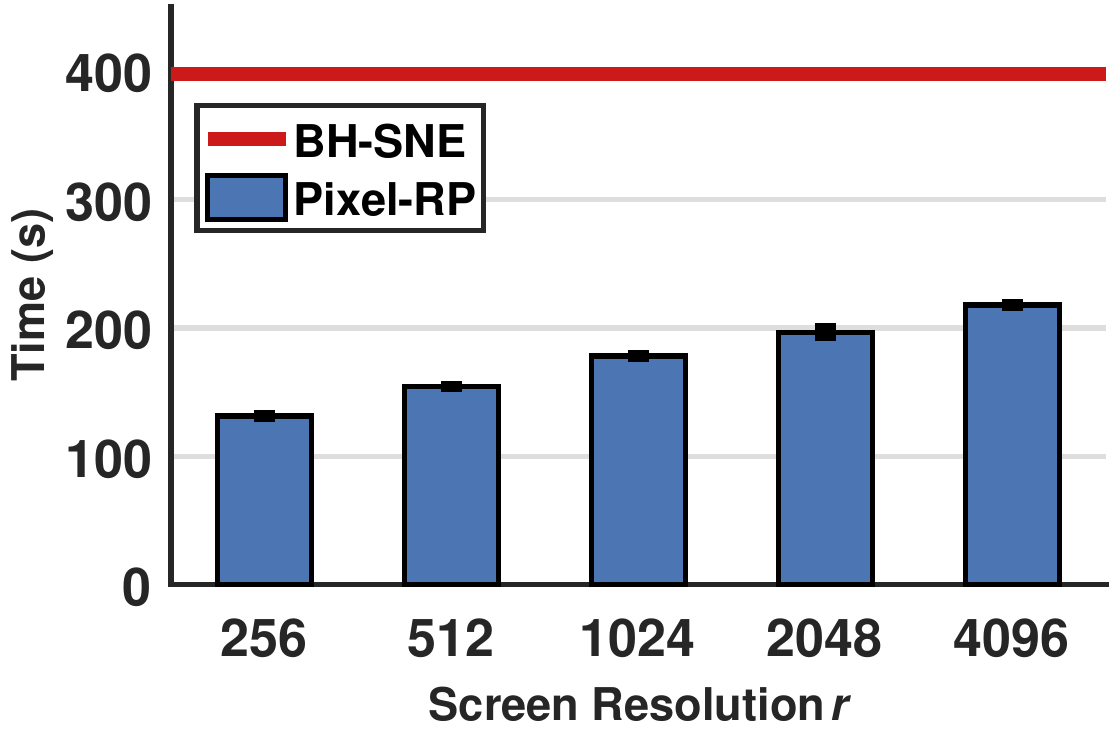}

\vspace{0in}
}\subfloat[50,000 data items]{\centering\includegraphics[width=0.5\columnwidth]{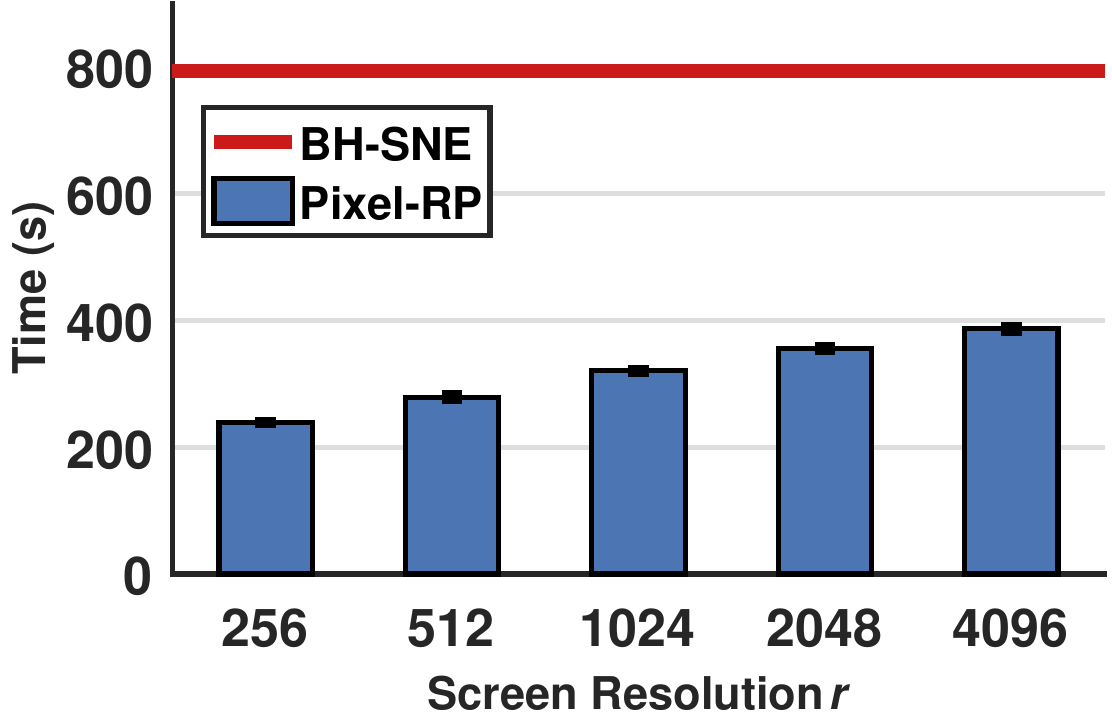}

\vspace{0in}
}\caption{Computing times of BH-SNE and PixelSNE-RP with the different screen
resolutions on MNIST dataset. The error bar represents the standard
deviation from five repetitions. }

\label{fig:mnist_bins}\vspace{-0in}
\end{figure}

\textbf{}%

\textbf{}%

\subsection{Computing Time Analysis}

\begin{figure*}[t]
\centering\subfloat[20News]{\includegraphics[width=0.2\textwidth]{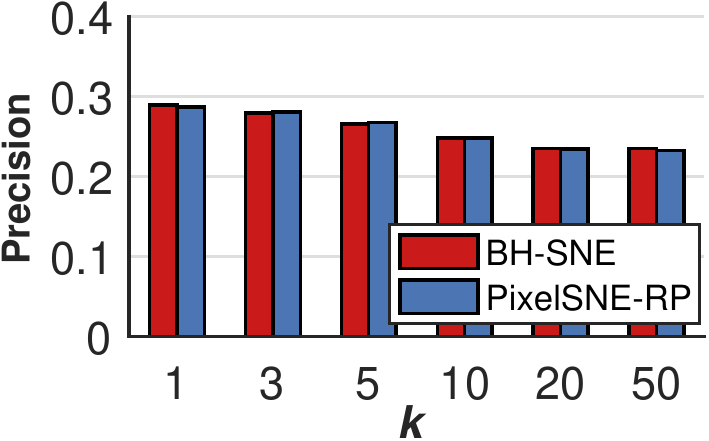} 

\vspace{0in}
}\subfloat[MNIST]{\includegraphics[width=0.2\textwidth]{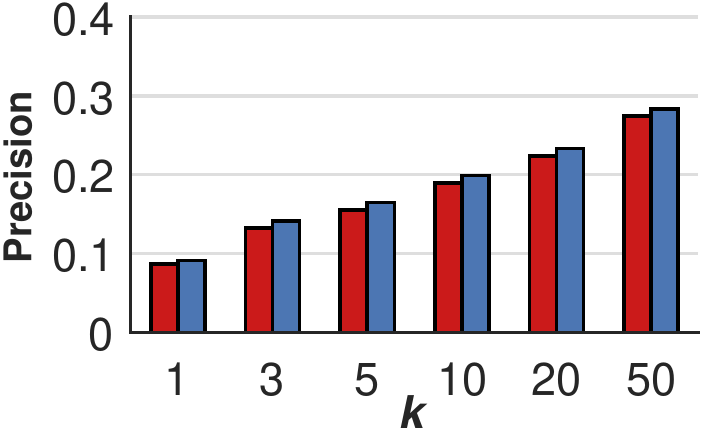} 

\vspace{0in}
}\subfloat[FacExp]{\includegraphics[width=0.2\textwidth]{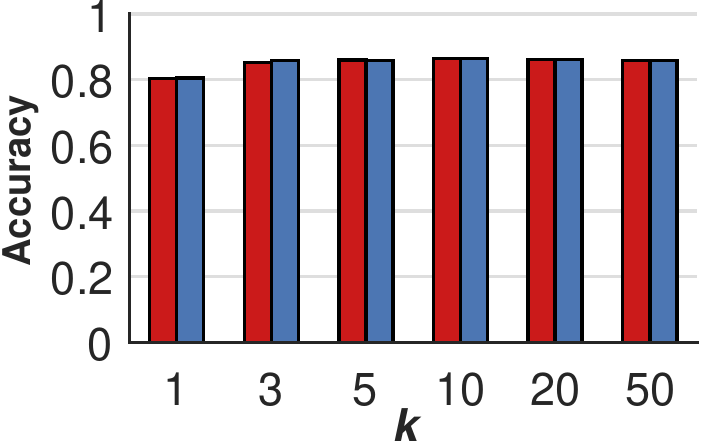} 

\vspace{0in}
}\subfloat[NewsAgg]{\includegraphics[width=0.2\textwidth]{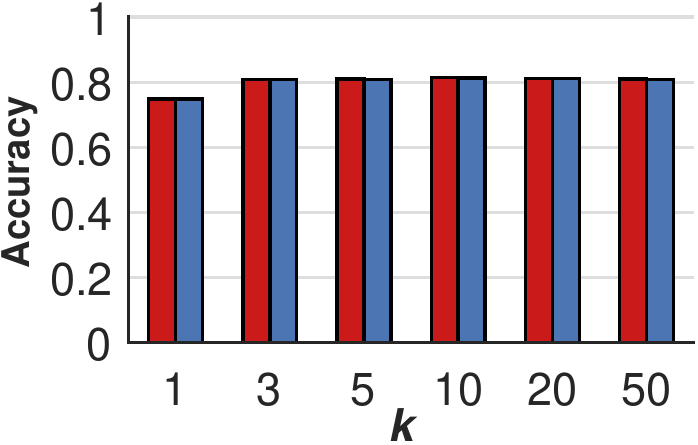} 

\vspace{0in}
}\subfloat[Yelp]{\includegraphics[width=0.2\textwidth]{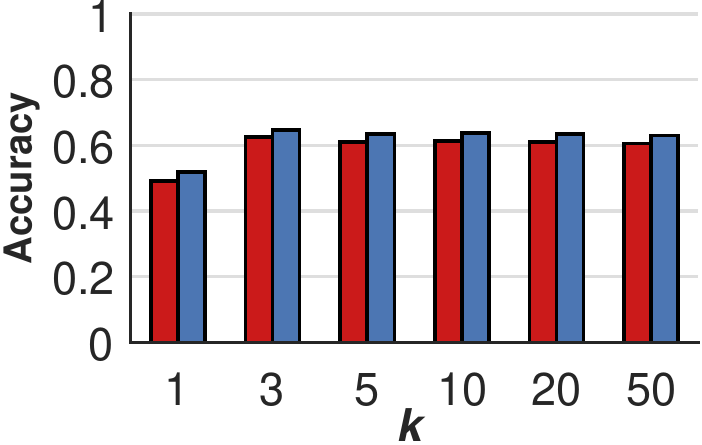} 

\vspace{0in}
}%
\caption{Embedding quality comparisons between BH-SNE and PixelSNE-RP. We report
the neighborhood precision (Eq.~(\ref{eq:nei_prec})) for (a) and
(b) and the $k$-NN classification accuracy for (c), (d), and (e). }

\label{fig:Quality_comparison}\vspace{-0in}
\end{figure*}

Table~\ref{tab: total_compute_time} shows the comparison of computing
times of various algorithms for different datasets. We report the
computing time of two sub-routines as well as their total times: (1)
constructing the original pairwise probability matrix $P$~(\textbf{\textit{P}}),
(2) optimizing low-dimensional coordinates~(\textbf{Coord}), and
(3) the total computing time~(\textbf{Total}). For fair comparisons,
the minor improvements presented in Section~\ref{subsec:minor_improvements}
are not applied to PixelSNE-VP nor to PixelSNE-RP. In addition, due
to its significant computing time, the computation time results of
t-SNE is excluded for large datasets such as Yelp or NewsAgg datasets.
{} 

\noindent\textbf{Comparison results}. In all cases, PixelSNE-VP and
PixelSNE-RP consistently outperform t-SNE and BH-SNE by a large margin.
For example, for Yelp dataset, BH-SNE took more than 18 hours while
PixelSNE-VP and PixelSNE-RP took 12 hours and less than 4 hours, respectively.
For the part of optimizing the low-dimensional coordinates~(\textbf{Coord}),
where we mainly applied our screen resolution-driven precision, PixelSNE-RP
and PixelSNE-VP both show the significant performance boost against
BH-SNE. For instance, PixelSNE-VP and PixelSNE-RP compute this part
more than three times faster than BH-SNE for NewsAgg dataset.%
{} When it comes to the part of constructing $P$~(\textbf{\textit{P}}),
as the size of the data gets larger, e.g., NewsAgg and Yelp datasets,
PixelSNE-RP runs significantly faster than PixelSNE-VP due to the
advantage of random projection tree adopted in PixelSNE (Section.~\ref{subsec:constructing_P}). 

\noindent\textbf{Scalability due to the data size}. Next, Fig.~\ref{fig:Total_time}
shows the computing times depending on the total number of data items,
sampled from NewsAgg dataset. %
As the data size gets larger, our PixelSNE-RP runs increasingly faster
than BH-SNE as well as t-SNE, which shows the promising scalability
of PixelSNE. 

\noindent\textbf{Effects of the precision parameter}. Fig.~\ref{fig:mnist_bins}
shows the computing time depending on the screen resolution parameter
$r$ for 30,000 and 50,000 random samples from MNIST dataset. Although
PixelSNE-RP is consistently faster than BH-SNE, it tends to run slowly
as $r$ increases. Nonetheless, our method still runs much faster,
e.g., around two-fold speedup compared to BH-SNE, even with a fairly
large values of $r$, i.e. 4,096, which can contain more than 16 million
pixels. %

\noindent\textbf{Effects of implementation improvements}. We compared
the computing time between PixelSNE-RP and improved PixelSNE-RP, which
adopts the efficient operation proposed in Section~\ref{subsec:minor_improvements}.
Table.~\ref{tab: bitwise} presents that the improved version consistently
brings around 4\% speed improvement compared to the original PixelSNE-RP
for all the datasets. Interestingly, our improvement also reduces
the variance of computing times. %

\begin{table}[th]
\caption{Computing times of PixelSNE-RP with and without the improvements shown
in Section~\ref{subsec:minor_improvements}). The numbers in the
parentheses represent the standard deviation from three repetitions. }
\label{tab: bitwise}

\centering%
\begin{tabular}{|c||c|c|c|c|c|}
\hline 
 & {\scriptsize{}20News} & {\scriptsize{}FecExp} & {\scriptsize{}MNIST} & {\scriptsize{}NewsAgg} & {\scriptsize{}Yelp}\tabularnewline
\hline 
\hline 
\multicolumn{1}{|c||}{{\footnotesize{}Improved PixelSNE-RP}} & {\scriptsize{}83.26s} & {\scriptsize{}134.43s} & {\scriptsize{}429.08s} & {\scriptsize{}3194.86s} & {\scriptsize{}13485.27s}\tabularnewline
{\footnotesize{}(Sec.~\ref{subsec:minor_improvements})} & {\scriptsize{}(0.19s)} & {\scriptsize{}(0.21s)} & {\scriptsize{}(9.05s)} & {\scriptsize{}(38.99s)} & {\scriptsize{}(51.1s)}\tabularnewline
\hline 
\multirow{2}{*}{{\footnotesize{}PixelSNE-RP}} & {\scriptsize{}87.7s} & {\scriptsize{}141.54s} & {\scriptsize{}446.88s} & {\scriptsize{}3370.36s} & {\scriptsize{}14025.12s}\tabularnewline
 & {\scriptsize{}(1.93s)} & {\scriptsize{}(4.99s)} & {\scriptsize{}(9.86s)} & {\scriptsize{}(150.04s)} & {\scriptsize{}(671.84s)}\tabularnewline
\hline 
\end{tabular}
\end{table}

\subsection{Embedding Quality Analysis}

\begin{table}[h]
\caption{Neighborhood precision (Eq.~(\ref{eq:nei_prec})) of PixelSNE-RP
on the 10,000 samples of MNIST dataset. The numbers in the parentheses
represent the standard deviation from ten repetitions.}
\label{tab: acc_bins_neigh}

\centering%
\begin{tabular}{|c||c|c|c|c|c|c|}
\hline 
\multirow{2}{*}{\textbf{\footnotesize{}$r$}} & \multicolumn{6}{c|}{\textbf{\scriptsize{}$k$ in $k$ nearest neighbors}}\tabularnewline
\cline{2-7} 
 & {\scriptsize{}1} & {\scriptsize{}3} & {\scriptsize{}5} & {\scriptsize{}10} & {\scriptsize{}20} & {\scriptsize{}30}\tabularnewline
\hline 
\hline 
\multirow{2}{*}{{\footnotesize{}512}} & {\scriptsize{}.2401} & {\scriptsize{}.3207} & {\scriptsize{}.3475} & {\scriptsize{}.3719} & {\scriptsize{}.3916} & {\scriptsize{}.4237}\tabularnewline
 & {\scriptsize{}(.0125)} & {\scriptsize{}(.0117)} & {\scriptsize{}(.0098)} & {\scriptsize{}(.0064)} & {\scriptsize{}(.0033)} & {\scriptsize{}(.0023)}\tabularnewline
\hline 
\multirow{2}{*}{{\footnotesize{}1024}} & {\scriptsize{}.2405} & {\scriptsize{}.3165} & {\scriptsize{}.3434} & {\scriptsize{}.3702} & {\scriptsize{}.3932} & {\scriptsize{}.4290}\tabularnewline
 & {\scriptsize{}(.0084)} & {\scriptsize{}(.0077)} & {\scriptsize{}(.0061)} & {\scriptsize{}(.0037)} & {\scriptsize{}(.0023)} & {\scriptsize{}(.0011)}\tabularnewline
\hline 
\multirow{2}{*}{{\footnotesize{}2048}} & {\scriptsize{}.2463} & {\scriptsize{}.3216} & {\scriptsize{}.3481} & {\scriptsize{}.3750} & {\scriptsize{}.3977} & {\scriptsize{}.4328}\tabularnewline
 & {\scriptsize{}(.0056)} & {\scriptsize{}(.0045)} & {\scriptsize{}(.0046)} & {\scriptsize{}(.0030)} & {\scriptsize{}(.0022)} & {\scriptsize{}(.0008)}\tabularnewline
\hline 
\multirow{2}{*}{{\footnotesize{}4096}} & {\scriptsize{}.2517} & {\scriptsize{}.3272} & {\scriptsize{}.3532} & {\scriptsize{}.3782} & {\scriptsize{}.4006} & {\scriptsize{}.4344}\tabularnewline
 & {\scriptsize{}(.0054)} & {\scriptsize{}(.0052)} & {\scriptsize{}(.0046)} & {\scriptsize{}(.0027)} & {\scriptsize{}(.0018)} & {\scriptsize{}(.0009)}\tabularnewline
\hline 
\end{tabular}
\end{table}

\textbf{Evaluation measures}. To analyze the embedding quality we
adopted the two following measures: 

\noindent\textbf{Neighborhood precision} measures how much the $k$
original nearest neighbors in a high-dimensional space are captured
in the $k$ nearest neighbors in the 2D embedding results. In detail,
let us denote the $k$ nearest neighbors of data $i$ in the high-dimensional
space and those in the low-dimensional (2D) space as $\mathcal{N}_{D}^{\left(k\right)}\left(i\right)$
and $\mathcal{N}_{d}^{\left(k\right)}\left(i\right)$, respectively.
The neighborhood precision is computed as 
\begin{equation}
\frac{1}{kN}\sum_{i=1}^{N}\left|\mathcal{N}_{D}^{\left(k\right)}\left(i\right)\cap\mathcal{N}_{d}^{\left(k\right)}\left(i\right)\right|\label{eq:nei_prec}
\end{equation}

\noindent\textbf{$k$-NN classification accuracy} measures the $k$-nearest
neighbor classification accuracy based on the 2D embedding results
along with their labels. 

\noindent\textbf{Method comparisons}. Fig.~\ref{fig:Quality_comparison}
shows the comparison results between PixelSNE-RP and BH-SNE depending
on different $k$ values. For the neighborhood precision measure,
PixelSNE-RP achieves the similar or even better performance compared
to the BH-SNE (Figs.~\ref{fig:Quality_comparison}(a) and (b)). We
conjecture the reason is because the random projection tree used in
PixelSNE-RP (Section \ref{subsec:constructing_P}) finds more accurate
nearest neighbors than the vantage-point tree used in BH-SNE~\cite{tang16largevis}.
For $k$-NN classification accuracy results shown in Figs.~\ref{fig:Quality_comparison}(c)-(e),
the performance gap between the two are small, indicating that our
method has a comparable quality of outputs to BH-SNE.

\begin{figure*}[t]
\centering\subfloat[PixelSNE-RP (3h 54m), $r=8192$]{\includegraphics[width=0.42\textwidth]{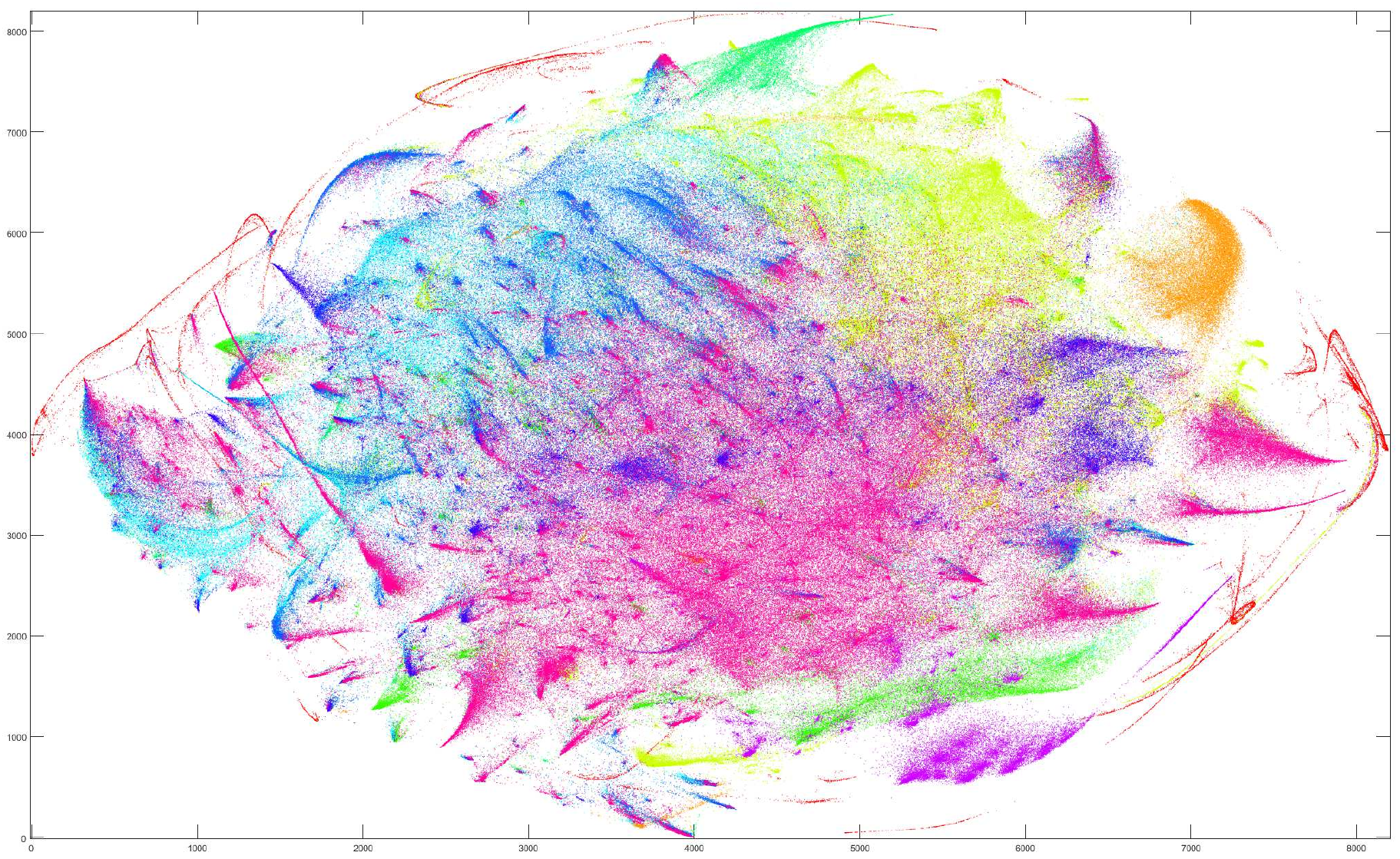} 

\vspace{0in}
}\hspace{1in}\subfloat[BH-SNE (18h 16m)]{\includegraphics[width=0.42\textwidth]{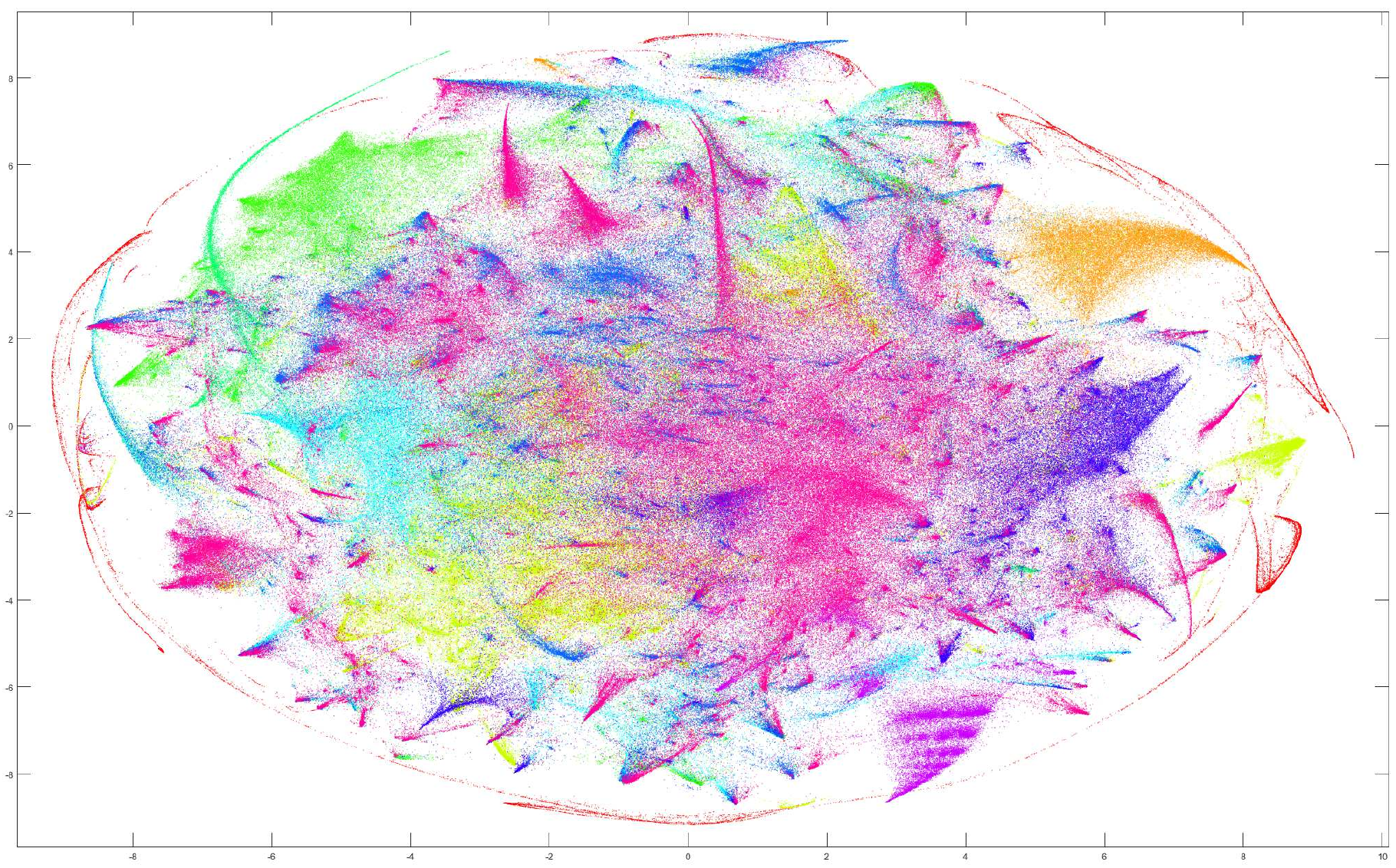} 

\vspace{0in}
}\caption{2D embedding of Yelp dataset. The numbers in parentheses indicate
the computing time.}

\label{fig:yelp}\vspace{-0in}
\end{figure*}

\begin{figure*}[t]
\centering\subfloat{\includegraphics[bb=0bp 0bp 842bp 327bp,width=0.7\paperwidth]{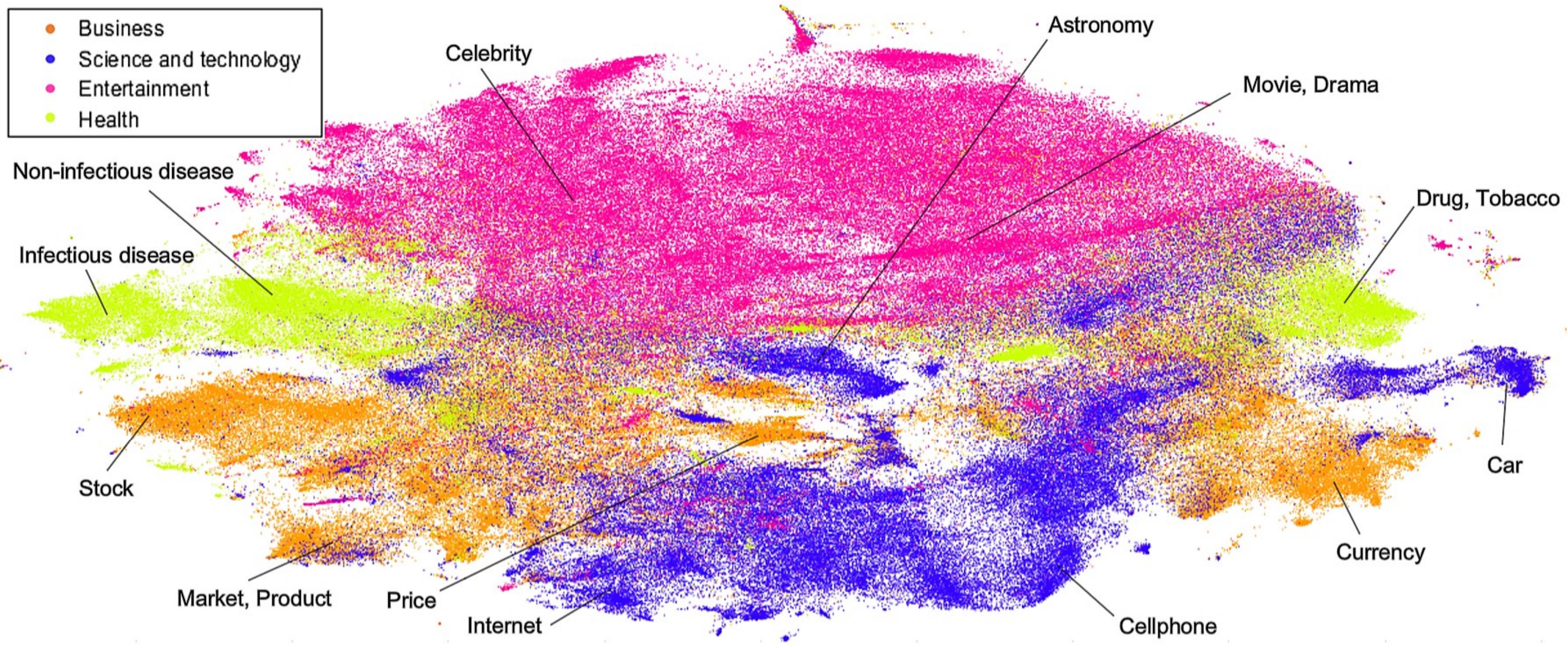} 

\vspace{0in}
} \caption{2D embedding of NewsAgg dataset generated by PixelSNE-RP. }

\label{fig:news}\vspace{-0in}
\end{figure*}

\begin{table}[h]
\caption{$k$-NN classification accuracy of PixelSNE-RP on the 10,000 samples
of MNIST dataset. The numbers in parentheses represent the standard
deviation from ten repetitions.}
\label{tab: acc_bins_knn}

\centering%
\begin{tabular}{|c||c|c|c|c|c|c|}
\hline 
\multirow{2}{*}{\textbf{\scriptsize{}$r$}} & \multicolumn{6}{c|}{\textbf{\scriptsize{}$k$ in $k$ nearest neighbors}}\tabularnewline
\cline{2-7} 
 & {\scriptsize{}1} & {\scriptsize{}3} & {\scriptsize{}5} & {\scriptsize{}10} & {\scriptsize{}20} & {\scriptsize{}30}\tabularnewline
\hline 
\hline 
\multirow{2}{*}{{\scriptsize{}512}} & {\scriptsize{}.9390} & {\scriptsize{}.9482} & {\scriptsize{}.9463} & {\scriptsize{}.9428} & {\scriptsize{}.9367} & {\scriptsize{}.9297}\tabularnewline
 & {\scriptsize{}(.0012)} & {\scriptsize{}(.0023)} & {\scriptsize{}(.0023)} & {\scriptsize{}(.0023)} & {\scriptsize{}(.0033)} & {\scriptsize{}(.0045)}\tabularnewline
\hline 
\multirow{2}{*}{{\scriptsize{}1024}} & {\scriptsize{}.9387} & {\scriptsize{}.9489} & {\scriptsize{}.9475} & {\scriptsize{}.9450} & {\scriptsize{}.9400} & {\scriptsize{}.9335}\tabularnewline
 & {\scriptsize{}(.0017)} & {\scriptsize{}(.0023)} & {\scriptsize{}(.0014)} & {\scriptsize{}(.0014)} & {\scriptsize{}(.0011)} & {\scriptsize{}(.0017)}\tabularnewline
\hline 
\multirow{2}{*}{{\scriptsize{}2048}} & {\scriptsize{}.9386} & {\scriptsize{}.9489} & {\scriptsize{}.9472} & {\scriptsize{} .9395} & {\scriptsize{}.9450} & {\scriptsize{}.9323}\tabularnewline
 & {\scriptsize{}(.0020)} & {\scriptsize{}(.0018)} & {\scriptsize{}(.0011)} & {\scriptsize{}(.0020)} & {\scriptsize{}(.0014)} & {\scriptsize{}(.0015)}\tabularnewline
\hline 
\multirow{2}{*}{{\scriptsize{}4096}} & {\scriptsize{}.9380} & {\scriptsize{}.9484} & {\scriptsize{} .9471 } & {\scriptsize{}.9437} & {\scriptsize{}.9382} & {\scriptsize{}.9324 }\tabularnewline
 & {\scriptsize{}(.0018)} & {\scriptsize{}(.0007)} & {\scriptsize{}(.0013)} & {\scriptsize{}(.0013)} & {\scriptsize{}(.0014)} & {\scriptsize{}(.0019)}\tabularnewline
\hline 
\end{tabular}
\end{table}

\noindent\textbf{Effects of the precision parameter}. Tables~\ref{tab: acc_bins_neigh}
and \ref{tab: acc_bins_knn} show the above two measures for PixelSNE-RP
with respect to different values of a screen resolution $r$. Unlike
the $k$-NN classification accuracy, which stays roughly the same
regardless of different values of $r$, the neighborhood precision
consistently increases as $r$ gets large. However, the gap is not
that significant.

\noindent\textbf{Cost value analysis}. Finally, Table~\ref{tab: cost_bins}
compares the cost function value (Eq.~(\ref{eq:cost_func})) after
convergence. Considering the baseline cost value of the random initialization,
both BH-SNE and PixelSNE-RP achieved a similar level of the algorithm
objective. Also, throughout all the values of $r$ tested, this value
remains almost the same between BH-SNE and PixelSNE-RP, which indicates
that the screen resolution-based precision of PixelSNE has minimal
to no impact in achieving the optimal cost value, which is consistent
with the results found in Tables~\ref{tab: acc_bins_neigh} and \ref{tab: acc_bins_knn}.%

\begin{table}[h]
\caption{Comparison of cost function values between BH-SNE and PixelSNE-RP.
The numbers in parentheses represent the standard deviation from ten
repetitions. }
\label{tab: cost_bins}

\centering%
\begin{tabular}{|c||c||c||c|c|c|c|}
\hline 
\multirow{2}{*}{} & \textbf{\scriptsize{}Random} & \multirow{2}{*}{\textbf{\scriptsize{}BH-SNE}} & \multicolumn{4}{c|}{\textbf{\scriptsize{}PixelSNE-RP (with $r$)}}\tabularnewline
\cline{4-7} 
 & \textbf{\scriptsize{}Coordinates} &  & {\scriptsize{}$512$} & {\scriptsize{}$1024$} & {\scriptsize{}$2048$} & {\scriptsize{}$4096$}\tabularnewline
\hline 
{\scriptsize{}Cost value} & {\scriptsize{}91.580} & {\scriptsize{}1.815} & {\scriptsize{}1.820} & {\scriptsize{}1.865} & {\scriptsize{}1.871} & {\scriptsize{}1.868}\tabularnewline
{\scriptsize{}(Eq.~\ref{eq:cost_func})} & {\scriptsize{}(.000)} & {\scriptsize{}(.0071)} & {\scriptsize{}(.0303)} & {\scriptsize{}(.0167)} & {\scriptsize{}(.0113)} & {\scriptsize{}(.0167)}\tabularnewline
\hline 
\end{tabular}
\end{table}

\subsection{Exploratory Analysis}

Finally, we presents visualization examples for Yelp and NewsAgg datasets.\footnote{High-resolution images and other comparison results are available
at \url{http://davian.korea.ac.kr/myfiles/jchoo/resource/kdd17pixelsne_appendix.pdf}. } Fig.~\ref{fig:yelp} presents the visualization generated from PixelSNE-RP
and BH-SNE on Yelp dataset. Different colors represent ten different
labels generated from topic modeling. Even though PixelSNE-RP ran
4.7 times faster than BH-SNE, the visualization results are comparable
to each other.

Fig.~\ref{fig:news} shows the visualization result on NewsAgg dataset
from PixelSNE-RP. A data point, which corresponds to a news headline,
is labeled with four different categories: Business, Science and Technology,
etc., denoted by different color. Although all the headlines are categorized
into only four subjects, Fig.~\ref{fig:news} revealed the additional
sub-category information by closely embedding news with the similar
topics. Each topic is connected with the relevant area as shown in
Fig.~\ref{fig:news}. For example, those news belonging to Health
category are roughly divided into three parts, as follow. The topical
region ``Infectious disease'' on the left side consists of the headline
with the keywords such as ``Ebola'', ``West Nile virus'' or ``Mers''
while the region connected with ``Non-infectious disease'' has headlines
with the keywords of ``Breast cancer'', ``Obesity'' or ``Alzheimer''.
The topical area of ``Drug'', ``Tobacco'' has news about health-related
life styles, e.g., ``Mid-life Drinking Problem Doubles Later Memory
Issues''. Note that the topical regions of ``Infectious disease''
and ``Non-infectious disease'' are closely located compared to the
other regions. 

Another interesting observation is that as long as news headlines
even from different categories share similar keywords, they are closely
located to each other. For example, the topic cluster of ``Market''
and ``Product'', on the lower-left region contains news from Business
and Science and technology. However, it turns out that the news from
Science and technology category has the headlines about the stock
price of the electronic company or the sales of the electronic products,
which is highly related to the headline about the stock market and
company performance from Business category. 

{}

\section{Conclusions}

\label{sec:conclusions} 

In this paper, we presented a novel idea of exploiting the screen
resolution-driven precision as the fundamental framework for significantly
accelerating the 2D embedding algorithms. Applying this idea to a
state-of-the-art method called Barnes-Hut SNE, we proposed a novel
approach called PixelSNE as a highly efficient alternative.

In the future, we plan to apply our framework to other advanced algorithms
such as LargeVis~\cite{tang16largevis}. We also plan to develop
a parallel distributed version of PixelSNE to further improve the
computational efficiency.

\bibliographystyle{ACM-Reference-Format}
\bibliography{Minjeong_Kim,JaegulChoo}

\end{document}